# Uncertainty Prediction for Machine Learning Models of Material Properties

Francesca Tavazza, Brian De Cost, Kamal Choudhary

Materials Science and Engineering Division, National Institute of Standards and Technology, Gaithersburg, MD, 20899, USA

**ABSTRACT**

Uncertainty quantification in Artificial Intelligence (AI)-based predictions of material properties is of immense importance for the success and reliability of AI applications in material science. While confidence intervals are commonly reported for machine learning (ML) models, prediction intervals, i.e., the evaluation of the uncertainty on each prediction, are seldomly available. In this work we compare 3 different approaches to obtain such individual uncertainty, testing them on 12 ML-physical properties. Specifically, we investigated using the Quantile loss function, machine learning the prediction intervals directly and using Gaussian Processes. We identify each approach's advantages and disadvantages and end up slightly favoring the modeling of the individual uncertainties directly, as it is the easiest to fit and, in most cases, minimizes over-and under-estimation of the predicted errors. All data for training and testing were taken from the publicly available JARVIS-DFT database, and the codes developed for computing the prediction intervals are available through JARVIS-Tools.

**Corresponding author:** Kamal Choudhary (kamal.choudhary@nist.gov)

# 1. INTRODUCTION

Artificial Intelligence (AI) approaches for Material Science have been explored for decades [1-9], but only in the last few years, they have become consistently successful across a wide variety of materials science tasks. While Material Science is still data-poor compared to the many other application areas of AI modeling, modern instruments and powerful computational facilities are finally able to produce enough data for successful application of Machine Learning (ML) in this area. The Materials Genome Initiative (MGI) has also contributed to this flourishing, by supporting the creation of large, publicly accessible databases, with very consistently computed data. With respect to atomistic computational results, regression models were first successfully applied to the evaluation of energetic-related properties, like formation energy[10-12], then expanded to predict elastic[13-15], electronic[5, 16, 17], optical[18] and many other physical properties. Computational parameters needed in simulations, not just physical quantities, can be predicted using regression models as well such as for k-points, as discussed in Ref[19]. JARVIS-ML[12] is an example of a publicly available suite of ML models for a variety of physical properties. Specifically, all its models apply to the prediction of properties for ideal, crystalline materials and have been trained on the density functional theory (DFT) data in the JARVIS-DFT database[20, 21]( https://jarvis.nist.gov/jarvisbdft/ )

ML methods are intrinsically statistical in nature, as the true underlying model form is not available to the ML model. However, quite surprisingly, during this period of explosion of AI applied to materials science, relatively little effort has been spent in evaluating its uncertainty, beyond assessing the average ability of the model. While there is a fair amount of discussion on how to evaluate uncertainty in ML/materials research[22-24], on how to estimate prediction intervals[25-30], and on how to collect data to improve ML models[31], the uncertainty on the individual prediction is often not rigorously evaluated or reported when machine learning material properties. Uncertainty Quantification (UQ) is an extremely important step in any experiment or computational assessments, as it determines how trustworthy the measured or computed data are. The same is true for the predictions of ML. Currently, the majority of ML/materials papers exclusively provide uncertainty evaluation on the average ability of ML model(s), providing quantities like the mean average error (MAE), the mean square error (MSE) or the root mean square error (RMSE). This approach fails to address the uncertainty/confidence of predictions for individual instances. For instance, when applying an ML model for formation energy to an unknown compound, we want the know if that specific material will be stable, and an evaluation of the average performance of the model over the entire dataset is not necessarily significant. Moreover, because the MSE-type stats make assumptions about the future data that we know we can't satisfy (independent, identically distributed data) in materials discovery and design settings, only providing these quantities as quality evaluators may be deceiving about the real capability of the ML model. The reason behind ML-UQ focusing mostly on population variables is that determining prediction intervals, i.e. the uncertainty on each specific prediction, can be computationally fairly expensive and, usually, it requires an additional computational effort beyond the training of the model. Gaussian Processes[32] are an exception to this rule, as the individual uncertainty is automatically determined when the model is fit. However, they have other limitations that often prevent them from being the approach of choice, as we will briefly discuss later in this work.

In this work we estimate prediction intervals using three straightforward and easy-to-implement approaches, one that allows to choose which population interval to be covered by the UQ (Quantile approach), the second, based on directly learning the error itself and applicable to any ML algorithm and the third, that automatically provides the uncertainty of each prediction when fitting the model (Gaussian Processes). Each method has its advantages and disadvantages, and they are the focus of the Results and discussion section. These methods are applied to energetic, mechanical, electronic and optical properties, and their results across properties are compared to each other. All the ML models in this work are trained and validated on the same DFT data, so that they can be compared in a completely consistent way. Specifically, for both training and validation, we use the DFT data contained in the JARVIS-DFT database[20, 21].

## 2. METHOD

All ML models discussed in this paper are regression models as implemented in LightGBM[33] in the case of Gradient Boosting Decision Tree (GBDT), or in Scikit-learn[34], for Gaussian Processes (GP) and Random Forest (RF). We use JARVIS-ML based classical force-field inspired descriptors (CFID)[12] when using GBDT, Gradient Boosting and Random Forest, and a subset of such descriptors when using GP. The CFID descriptors provide a complete set of structural and chemical features (1557 for each material) and CFID based models have been successfully used to develop more than 25 high-accurate ML property prediction models[12]. All the ML models in this work are trained and validated on the DFT data contained in the JARVIS-DFT database[20, 21]. This allows for a consistent comparison between methodologies.

To determine prediction intervals, the first methodology employed in this work estimates the error through the evaluation of an upper and a lower prediction bound for the quantity under examination. In the rest of the paper we will call this approach "Quantile" as it is based on using the quantile loss function. Machine learning models are often fit through the minimization of a loss function[22-24] as such a function quantifies the errors the model makes on the training data. There are many options for such loss functions, depending on factors like the distribution of the data, the choice of machine learning algorithm, the quality of the data etc. Some loss functions penalize low and high errors equally, others don't. The least-square loss function (*least square*, aka *MSE*) is a commonly used regression loss among those that penalize low and high errors equally, meaning that a positive and a negative error of the same amount incur the same penalty. Obviously, errors of different amount are penalized differently. Specifically, when using *MSE*

$$MSE = \frac{1}{N} * \sum_{i=1}^{N}(y_i - y_i^P)^2 \qquad (1)$$

optimal parameters are determined by minimizing the sum of squared residuals. Here $y_i^P$ are the values predicted for the target values $y_i$. In other words, *MSE* optimizes for the mean of the distribution.

Mean Absolute Error (MAE) is another example of loss function that penalizes low and high errors equally. MAE is less sensitive to outliers than MSE, so, depending on the problem, it may be a better choice. Other loss functions penalize low and high errors unequally, therefore allowing to optimize by percentiles. The quantile loss function (*quantile*) is part of this latter group, and what percentile it optimizes for depends on the choice of the quantile parameter α [22]:

$$Quantile\ loss = \begin{cases} \alpha * |y_i - y_i^P| & if\ (y_i - y_i^P) \geq 0 \\ (\alpha - 1) * |y_i - y_i^P| & if\ (y_i - y_i^P) < 0 \end{cases} \qquad (2)$$

where α is between 0 and 1. Such a property is the base of the first method we use to estimate the prediction intervals, therefore the name "Quantile" for this approach. For each physical quantity, we are interested in, we will independently optimize 3 ML models, one for $\alpha_{upper}$, one for $\alpha_{lower}$ and one for $\alpha_{mid}$, where $\alpha_{mid}=0.5$ (*i.e.,* the median). In the rest of the paper we will refer to these models as UPPER, LOWER and MID, respectively. The prediction interval for each predicted data is given by

$$PredInt(y_i^P) = |y_i^{upper} - y_i^{lower}| \qquad (3)$$

where $y_i^{upper}$ ($y_i^{lower}$) are the values predicted for $y_i$ using $\alpha_{upper}$ ($\alpha_{lower}$) and $y_i^P$ is the value predicted for $y_i$ using $\alpha_{mid}$. This will be referred to as the "predicted value" in the rest of the paper.

A nice advantage of this method is that the population interval to be covered is arbitrary. Most of the results discussed in this work have been obtained using $\alpha_{upper} = 0.84$ and $\alpha_{lower} = 0.14$ to construct the upper and lower bounds, so that our prediction interval would ideally cover 68 % of the population, corresponding to 1 standard deviation under the assumption of Gaussian error distribution. This makes for an easy comparison with the other methodologies examined in this paper. It is important to note, however, that the coverage of 68 % of the population (in case of $\alpha_{upper} = 0.84$ and $\alpha_{lower} = 0.14$) would only occur for perfect fits, and that, in general, the direct count of how often the reference value falls within the UPPER-LOWER interval gives a better idea of the meaning of the uncertainty. We will refer to this quantity as "in-bounds" in the rest of the paper.

Algorithm-wise, we utilize the Gradient Boosting Decision Tree (GBDT) regression for the Quantile approach, because it allows optimization of an arbitrary differentiable loss function, so that the MID model could be run using either quantile with $\alpha_{mid}$=0.5 (*i.e.,* optimizing for the median) or MSE (i.e., optimizing for the mean). Another advantage of GBDT is that it provides some degree of interpretability through feature importance ranking, a property we took advantage of when choosing descriptors for Gaussian Processes. As a standard practice, we use a 90 % to 10 % train-test split[12, 15, 17]. The 10 % independent test set is never used in the hyperparameter optimization or model training, so that when the model is evaluated on it, we obtain unbiased performance metrics.

The second approach we use to determine prediction intervals uses a machine learning model to predict the errors directly. This means that it requires the fitting of 2 ML models for each physical quantity under examination: one to predict the actual value of the property ("base" model) and one to determine the prediction intervals ("error" model). Consequently, training the models requires splitting the data in 3 groups: one to fit the base model, one to fit the error model, which can also be used to validate the base model, and the last one to validate the error model. Obviously, the third set can be used to validate the base model as well. Requiring splitting the data into three sets may be a substantial disadvantage when the amount of available data is small to begin with. We chose to work with a 45-45-10 split, so that the number of samples in the third set of data matches the number of data used to validate the other two methodologies. This makes the comparisons between approaches more consistent. In the rest of the paper this approach will be referred to as "3split", because of its requirement of splitting the available data in 3 groups.

To model the error using ML, we explored two approaches: one where, for each datapoint, we predicted the absolute value of the difference between exact and predicted value (Error_AD), and the other where we predicted the square of such a difference (Error_SD), instead:

$$Error\_AD_i = \left| y_i^{Observed} - y_i^{Predicted} \right| \qquad (4)$$

$$Error\_SD_i = \left( y_i^{Observed} - y_i^{Predicted} \right)^2 \qquad (5)$$

These two options were inspired by the comparison between mean absolute error (MAE) and mean square error (MSE) loss functions (L1 versus L2): using the squared of the differences is easier to optimize, but using the absolute value of such differences is more robust to outliers. They will be indicated as "3split-L1" (4) and "3split-L2" (5) in the rest of the paper.

An important advantage of this second approach is that it doesn't require a specific loss function, which means that it can be used with any regression model. We used the GBDT regressor, for a direct comparison to the Quantile approach, and the Random Forest Regressor, as our preliminary tests showed it to be very effective even without hyperparameter optimization. As for the Quantile approach, we computed the in-bounds percentage for each property model, to establish the meaning of the determined uncertainty.

The third approach we investigated uses Gaussian Processes, so that individual uncertainties are automatically determined as the model is trained[32]. Fitting a Gaussian process (GP) starts with the choice of the covariance function, also called *kernel*, so that the prior distribution is defined. A prior is an infinite-dimensional multivariate Gaussian distribution completely specified by a mean function, m(x), and a covariance function, k(x, x'):

$$f(x) \sim GP(m(x), k(x, x')) \qquad (6)$$

In other words, the prior does not depend on the training data but it's a way to incorporate prior knowledge about the functions. The next step is determining the posterior distribution (predictive), which is obtained by limiting to functions going through the training data. Because we have a Gaussian process prior, the posterior is a Gaussian process as well: it is completely described by mean and covariance and automatically provides the UQ estimate. The training of a GP model involves the model selection (i.e. the choice of the kernel function) and the optimization of the kernel hyperparameters. In this work, all fitting parameters ($\lambda_i$ for $i=1,..N$, $l$, $l'$, $l''$, $p$, $\alpha$, noise, $c_i$, $i$=1,2,3) are optimized using the Broyden–Fletcher–Goldfarb–Shanno[35] (L-BGFS-B) algorithm, as implemented in scipy.optimize.minimize.

The main advantage of using GP is the fact that the uncertainty on each prediction is determined together with the prediction value itself, without requiring any extra effort. Gaussian Processes also are extremely effective with a smaller set of data and allow to easily incorporate data into the model. However, GP have disadvantages as well, especially the fact that they lose efficiency in high dimensional spaces, i.e., if the number of features exceeds a few dozens or for large datasets. They are computationally memory intensive for large datasets as well.

Lastly, like in the cases of Quantile and "3split", the meaning of the error on the individual prediction comes from knowing how often the observed value (the DFT data in this work) falls into the prediction interval. Because the posterior distribution of a Gaussian Process is a Gaussian itself, the meaning of the uncertainty on each data should be straightforward. However, as in the case of Quantile, the fitting is never perfect, so it's important to compute its "inbounds" count on the validation set to establish its meaning for each fitted GP model.

## 3. RESULTS

All results discussed in this session refer to test data. In all cases we performed a 90-10 split, so the test data were 10 % of the whole dataset.

### 3.1 Prediction intervals: Quantile approach

When analyzing prediction intervals obtained using the Quantile approach and equation (3), the first characteristic that appears evident is their asymmetry with respect to the predicted value $y_i^P$, as shown in Fig 1 a) for a sub-set of formation energy data. However, for plotting purposes and to be able to directly compare Quantile-given errors to prediction intervals obtained using the other methodologies, in the rest of the paper we will define a *quantile error* as

$$err_{quantile} = PredInt(y_i^P)/2 \qquad (7)$$

where $PredInt(y_i^P)$ is given by (3), and we will investigate $y_i^P \pm err_{quantile}$.

This approach is exemplified in Figure 1 b), where same predicted data as in Figure 1 a) are displayed, and the size of their predicted intervals are unchanged as well, but now the error bar is given by $err_{quantile}$.

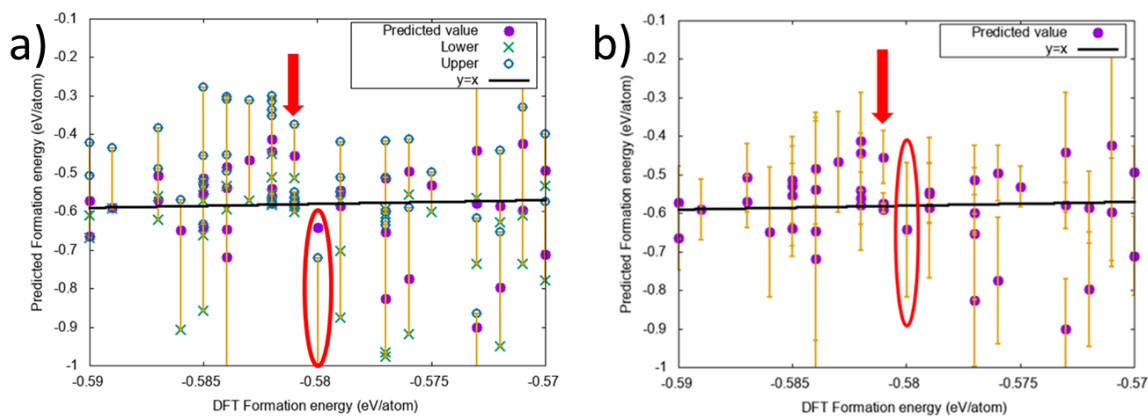

*Figure 1 a) Quantile prediction intervals for formation energy (enlargement). The upper and lower predictions determine the size of the prediction interval while the "mid" prediction determines the predicted value $y^P$. The y=x line shows where the predicted value should be to match the DFT data. The arrow points out a case where the predicted intervals is not large enough to include the observed (DFT) data. The elliptical enclosure shows an example of predicted value outside the prediction interval. b) Formation energy data (enlargement) with error bar given by $err_{quantile} = PredInt(y_i^P)/2$. The y=x line indicates where the predicted values should be to match the DFT data.*

Figures 1 a) also showcase possible problems that come with using this approach: the arrow points out a case where the predicted intervals is not large enough to include the observed (DFT) data, while the elliptical enclosure shows an example of predicted value outside the prediction interval. In general, $y_i^P$ may not fall between $y_i^{upper}$ and $y_i^{lower}$ because all 3 values are predicted independently, using models with different optimized hyperparameters, although identical training and validation sets. The same "problem" points are pointed out in Fig 1 b), but, now, because of the symmetric definition of the error, the second type of problem is eliminated by construction (the ellipse case, in the figure). The possibility of having the observed value (DFT data, indicated by the y=x line in Figure 1) outside the prediction interval cannot be eliminated, and, therefore, it is necessary to compute the in-bounds percentage for each model, using the validation set.

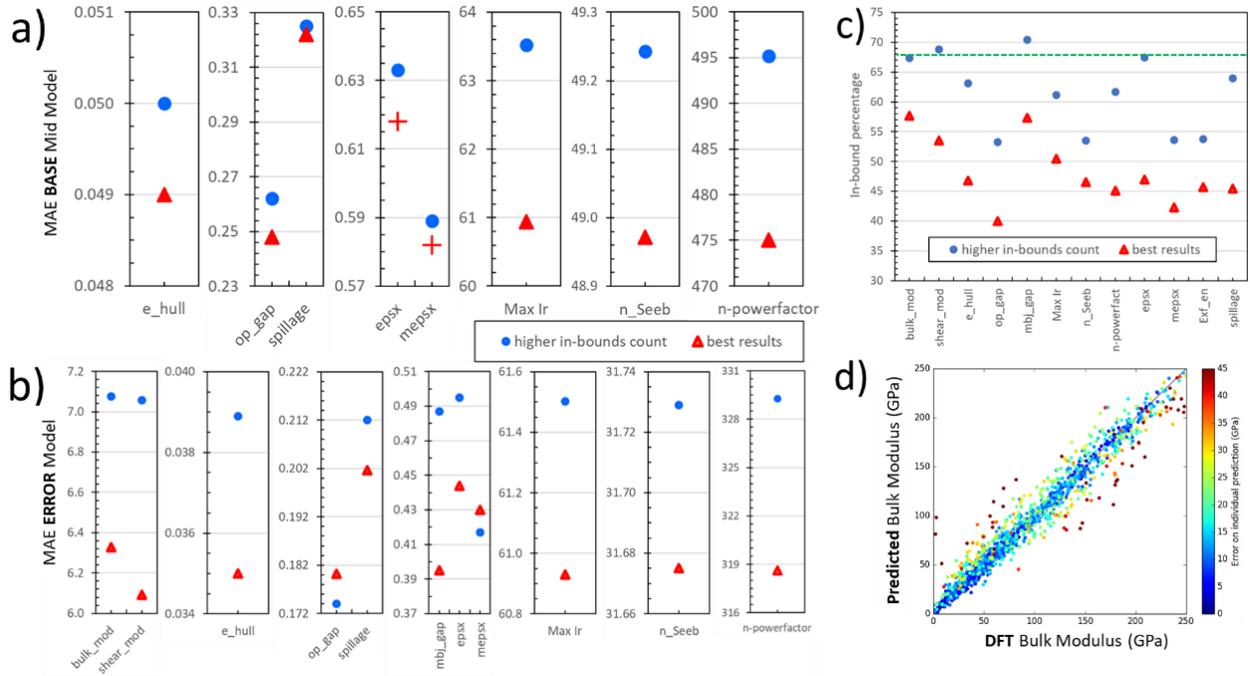

*Figure 2 Comparison between lowest MAE models ("best results") and models with an in-bound count closer to 68 % ("higher in-bounds count"), using Quantile. MAEs for Base Mid models are shown in a), corresponding MAEs for Error models in b) and in-bounds counts in c). An example of the prediction distribution is given in d), for bulk modulus in the "best results" case. The color of each point represents the size of its prediction interval. The dashed line in c) indicates the 68 % in-bound counts that would give easy interpretability to the error bars. Each MAE is in the units of the corresponding property (bulk_mod, shear_mod = bulk and shear modulus, respectively (GPa), e_hull = energy above the convex hull (eV), op_gap, mbj_gap = bandgap using OPTB88vdW, MBJ data, respectively (eV), spillage: no units, epsx(mepsx) = refractive index along x using OPTB88vdW, MBJ data, respectively (no-units), Max_ir = maximum infrared frequency (cm$^{-1}$), n_Seeb = Seebeck coefficient for electrons ($\mu$V/K)).*

Another point worth making, when discussing how to use the Quantile approach for uncertainty evaluation, is that the predicted value $y_i^P$ can be determined using the least square loss function (*MSE*), instead of the quantile loss with $\alpha_{mid}$ = 0.5. The use of *MSE* corresponds to driving the predictions towards the mean of the distribution instead of to its median, and it is the approach followed in this work.

We are now ready to examine results obtained implementing the Quantile approach. In Fig. 2 we compare two sets of models for each physical quantity of interest. The first set was obtained independently optimizing the hyperparameters of LOWER, MID and UPPER, to achieve the lowest possible MAE in each case, and it is indicated as "best results" in the figure. Figure 2c) shows that the in-bound percentages for "best results" are always lower than the wished-for 68 % (dashed line) and as low as 40 % in some cases. By choosing UPPER and LOWER models with slightly larger MAEs, it's possible to increase the in-bound counts. The new results are labelled "higher in-bounds count" and, while 68 % is not always obtained, the in-bound counts are closer to it than before. MAEs for Base Mid models are displayed in a), but only for those quantities for which the property prediction changed between the two cases. Error MAEs for all quantities are reported in b).

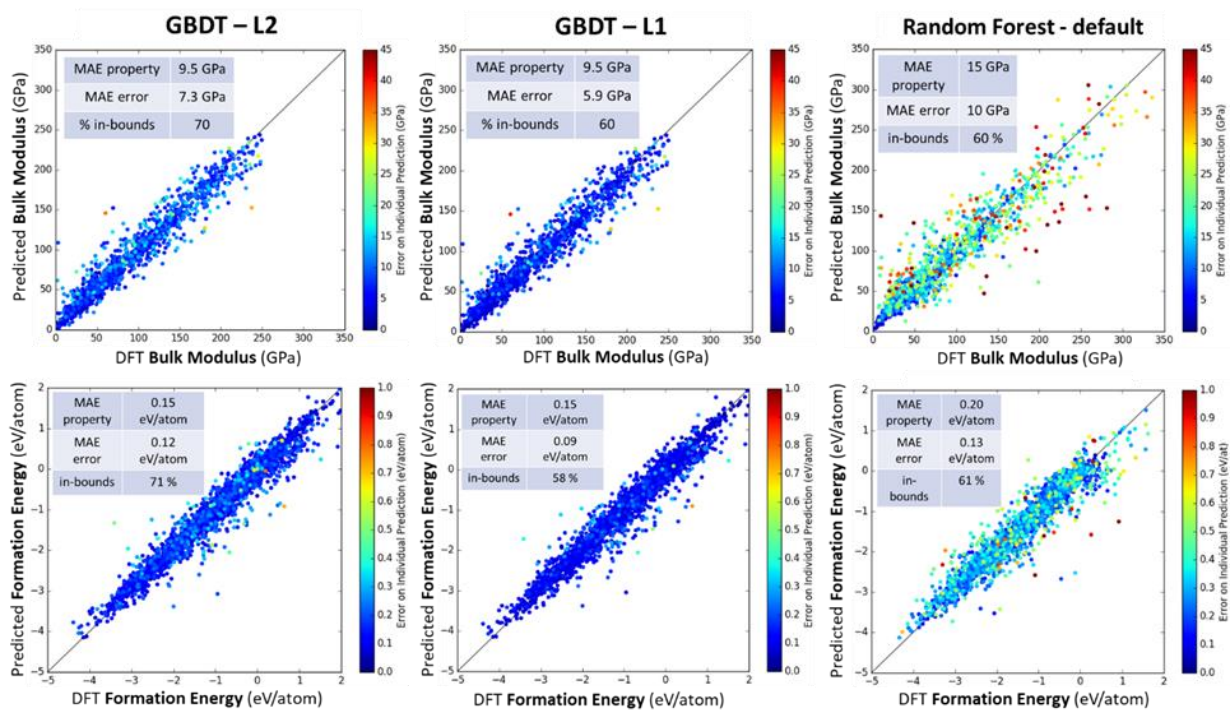

*Figure 3 Property and error predictions using different algorithms: Gradient Boosting decision tree (GBDT) with optimized hyperparameters and error defined either as the square or as the absolute value of the difference between observed and predicted value (L2 or L1, respectively), and Random Forest with default parameters. The error on each datapoint is indicated by its color. Results are given for bulk modulus (top row) and formation energy (bottom row). For each case, MAEs for both property and error are reported, as well as their in-bound counts.*

## 3.2 Prediction intervals: ML modeling of the error directly

As the main advantage of modeling the error directly is that any ML algorithm can be used, in Fig. 3 we compare optimized Gradient Boosting decision tree (GBDT) results to Random Forest (RF) ones, where RF was run using default parameters only. The reason for such a comparison is the fact that fitting ML models and, especially, optimizing their hyperparameters, is a computationally expensive endeavor. Adding UQ on individual predictions necessarily increases such computational cost, even significantly if ad-hoc hyperparameter optimization is needed. A possible work-around is to use avoid hyperparameters fitting, at least for the UQ part. Such an approach is feasible only if an algorithm is available, that produce acceptable MAE using default hyperparameters. Our investigation identified Random Forest as such an algorithm, with default parameters as in its scikit-learn implementation. While default-RF results are clearly worse than those from optimized GBDT, Figure 3 shows that their in-bound percentage is like the one from GBDT 3split-L1 and that that the error modeling is at least mostly qualitatively correct: blue points are on or near the diagonal, red and brown point are away from it. However, the presence of more light blue or green points on or very near the diagonal in RF results than in GBDT ones tells us that RF has a more case of overestimation than GBDT.

Figure 3 also provides visual comparison between 3split-L1 and 3split-L2 results: in both the bulk modulus and the formation energy cases, 3split-L1 has lower error MAE but also lower inbounds percentage than 3split-L2. Also, less light blue-colored points are seen in its plots, especially over or near the y=x line that indicates perfect fit, possibly indicating not as much overestimation. Conversely, more dark-blue points are seen away from the diagonal for 3split-L1 than for 3split-L2, indicating occasional underestimation of the errors. Finding lower error MAE and lower in-bounds for 3split-L1 than for 3split-L2 is a trend common to all physical quantities that we examined. More on how 3split-L1 and 3split-L2 models compare to each other and to Quantile and Gaussian Processes results will be covered in the Discussion section (Section 3.4). Lastly, it is well known that prediction intervals are always wider than confidence intervals [4], so it's no major concern that the errors are in many cases larger than the corresponding MAEs.

### 3.3 Prediction intervals: Gaussian Processes

In this work, we limited the investigation of Gaussian Processes (GP) to 6 quantities: formation energy, exfoliation energy, bulk modulus, bandgap obtained using the Tran-Blaha modified Becke-Johnson approach (MBJ bandgap), maximum infrared mode and spillage. Such quantities were chosen to cover a variety of different physical properties, spanning from energetics to elastic, electronic, optical and topological, while only probing some of the smaller datasets available to us, as GP becomes inefficient with larger sets. A major exception to this criterium is the inclusion of formation energy, as it comes with one of our largest datasets. It was important to add it, though, because formation energy is a quantity systematically investigated in ML works focused on nanoscale material science, and, therefore, analyzing its single prediction UQ is of wide-spread interest.

The choice of kernel is crucial to successfully fit GP. The specific kernel used in this work depends on the property of interest, but they all were a combination of two or more of the following kernel functions[36]:

$$K(x, x') = c1 * RBF(\lambda_i, i = 1, .., N) + c2 * RQ(\alpha, l) + c3 * RBF(l') * ExpSineSquared(l'', p) + c4 * WhiteNoise(noise) \qquad (8)$$

where:

- RBF (Radial-basis function kernel) is of the form $\exp(-\frac{d(x,x')}{2\lambda^2})$, where $d(x,x')$ is the Euclidean distance between x and x', and $\lambda$ is a vector of length scale parameters with as many dimensions as the number of descriptors (N).

- RQ (Rational quadratic kernel) is of the form $(1 + \frac{d(x,x')^2}{2\alpha l^2})^{-\alpha}$ and it is parameterized by a length-scale parameter $l$ and a scale mixture parameter $\alpha$.

- ExpSineSquared kernel is of the form $\exp\left(-\frac{2\sin^2(\pi d(x,x')/p)}{l''^2}\right)$, where l'' is the length-scale parameter and p is a periodicity parameter.

- WhiteNoise kernel is equal to the noise parameter "noise" for $x=x'$, it's null otherwise.

- $c_1$, $c_2$, $c_3$ and $c_4$ are fitting constants and can be set to zero.

Through the RBF term we include long term, smooth rising behavior, while smaller, medium term irregularities are made possible through the RQ component, in case periodic trends need to be considered. As the name hints, the WhiteNoise component of the kernel accounts for white noise

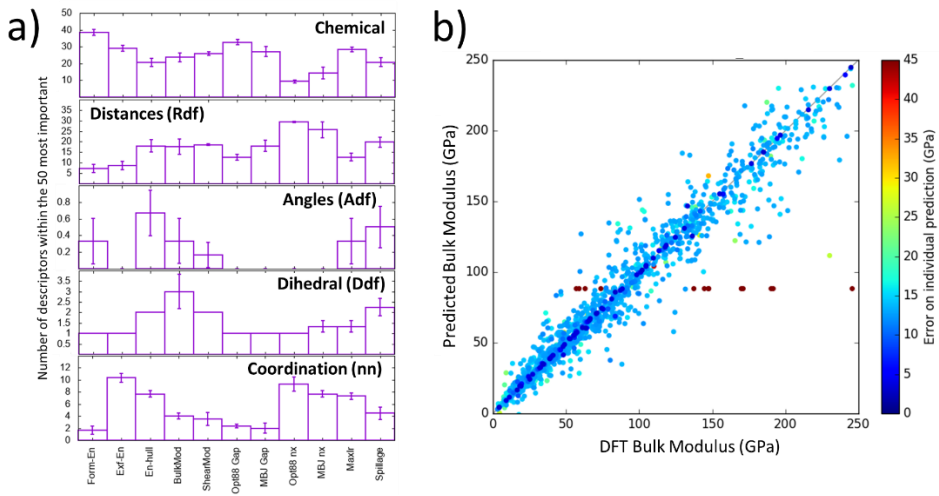

*Figure 4 a) Property-dependent descriptor importance analysis: for each property, the 50 most important descriptors are classified into their type (chemical, distance-related (Rdf = radial distribution function), angle-related (Adf = angular distribution function), dihedral angle related (Ddf = dihedral distribution function) and coordination-related (number of neighbors)). b) Prediction versus observed (DFT) data for bulk modulus. The color gives the predicted error on each individual prediction, using the same scale as in Figure 2 and 3, to facilitate comparisons.*

on the data. As our descriptors are very different in nature, going from chemical ones to structural and coordination-related ones, we found the importance to allow different length scale parameters for each of them, at least in the RDF term. All fitting parameters ($\lambda_i$ for i=1,..N, l, l', l'', p, $\alpha$, noise, $c_i$, i=1,2,3) are optimized using the Broyden–Fletcher–Goldfarb–Shanno (L-BGFS-B) algorithm[37].

CFID descriptors provide a complete set of structural and chemical features through 1557 descriptors for each material. Although 1557 is not an unreasonable number of descriptors/features for most ML models, they are too many to be effectively dealt with using Gaussian Processes. We used the "feature importance" capability of Gradient Boosting methodologies to reduce their number to something manageable, between 15 and 30, depending on the property. In Figure 4 a) data leading to our choice of which descriptors to use for each property are shown. Out of the 50 most important descriptors, the majority are chemical descriptors, for all properties but optical. After the chemical ones, either those distance-related or coordination-related are the most common ones. The error bars come from averaging importance results obtained in the three most successful GBDT models, for each property. There is some variability, among runs, on which descriptor, in each category, is among the 50 most important. Therefore, the intersection of the 50 most important descriptors between the 3 best models, for each property, gave us a manageable number of descriptors to use when fitting Gaussian Processes.

Figure 4 b) shows our results in the case of bulk modulus. Compared to Quantile (Figure 2) and "3split" (Figure 3) results, GP predicts less "very accurate" errors (dark blue, less than 2 GPa error), many more "accurate" ones (medium blue, (2 to 10) GPa error) and almost none with error larger than 17 GPa (green or above). This behavior, a reduced dispersion in the predicted error, occurs consistently in all 6 investigated properties, and always corresponds to a very high in-bounds percentage (usually around 80 %).

## 3.4 Prediction intervals: Discussion

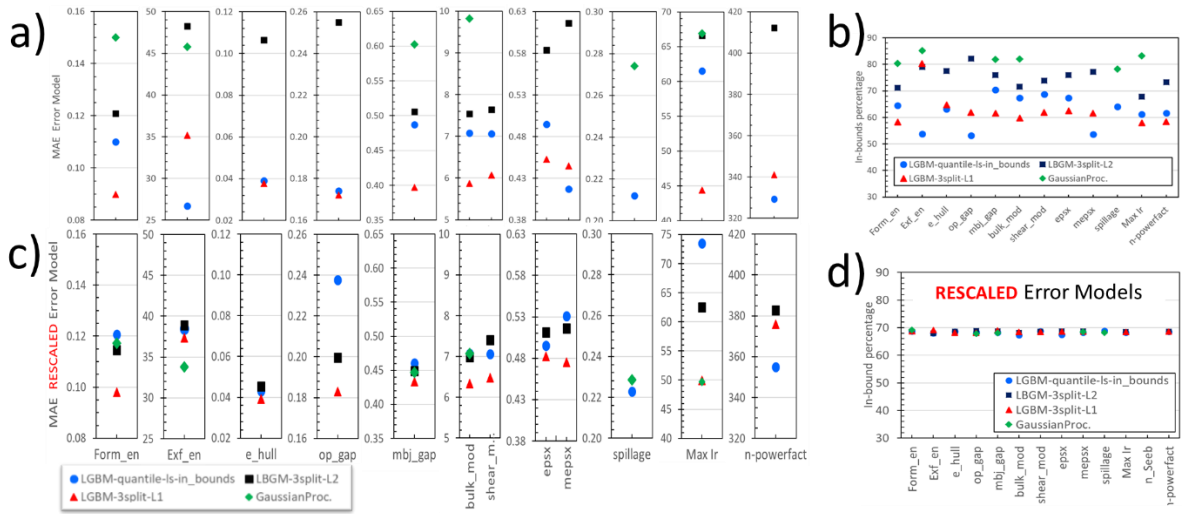

*Figure 5 Population variables for the 4 error models investigated in this work: MAEs in a) and inbounds percentages in b). Same quantities for rescaled error models in c) and d). As the inbounds percentages become very similar among models, the corresponding MAEs get closer as well, with 3split-L1 being the lowest in most properties. To facilitate comparison, we kept the y-axis scale the same between the error model as-fitted cases (a) and the rescaled ones (c). Each MAE is in the units of the corresponding property and details on the properties and their units are given in the caption of Figure 2.*

After having explored individual results from each of our modeling approaches, we are finally ready to quantitatively compare them across all the properties. Fig. 5a) shows error MAEs results, for all 4 approaches, having chosen models with low MAEs and inbounds percentages closest to 68 %. Having comparable inbounds count is necessary to get an accurate MAEs comparison. However, as displayed in Fig. 5c), no matter how much time and focus we put in the fits, we couldn't obtain models with exactly 68 % inbounds percentages (one standard deviation). Again, fitting for lower MAEs and higher inbounds counts at the same time is not straightforward, as the hyperparameter search is driven only by the minimization of the loss function. Fig. 5c) shows that GP inbounds count is systematically around 80 %, 3split-L2 between 70 % and 80 %, while 3split-L1 is consistently around 60 %. The only exception is Exfoliation Energy, which is the quantity with an incredibly small dataset (only about 600 datapoints), so its results are often unpredictable. Quantile has the largest spread in its inbounds percentages, from just above 50 % to 70 %. In terms of MAEs for the error models, Quantile and 3split-L1 are consistently lower than 3split-L2 and GP. The systematic difference between 3split-L1 and 3split-L2 hints to the importance of how the error is modelled (in 3split-L1 the absolute value of the error is machine-learned directly, while in 3split-L2 it is the absolute value of the error to be machine-learned directly). All MAEs and inbounds counts discussed here (error models) are given in Table S1 in the supplementary information (SI), together with the corresponding MAEs for the base models (i.e., the models predicting the physical quantities themselves).

As an accurate comparison of MAEs for the four methodologies is only possible for very similar inbounds counts, we rescaled the prediction intervals, as to obtain an inbounds percentage of 68 %. This was accomplished by multiplying all validation prediction intervals, for each physical property and error-modeling approach, by a fitted factor, chosen so that the inbounds percentage is 68 % or as close to it as possible. Such factors are approach and property-dependent and they are listed in Table S2 in the supplementary information. The error model MAEs are then recomputed using the scaled, validation sample values. Results are given in Fig. 5. By construction, all rescaled inbounds counts are very close to 68 % (Fig. 5d). The corresponding MAEs (Fig. 5c) are closer together than for the raw (not rescaled) case (Fig. 5a), highlighting how important it is to compare prediction-interval-determining methodologies only when the inbounds counts are similar. For certain properties, all models give essentially the same MAE (MBJ-band gap, for instance), while in other cases, like for the maximum predicted infrared mode (Max IR), there is still a substantial difference. Overall, it is 3split-L1 to achieve the lowest error MAEs across the most properties, followed by GP. As the "3split" models are the easiest/fastest to fit, this is a nice finding. Quantile doesn't always work as well as it could have been expected: in the case of two properties (OPTB88vdW(OPT)-bandgap and Max IR), it gives the highest MAEs for the error model.

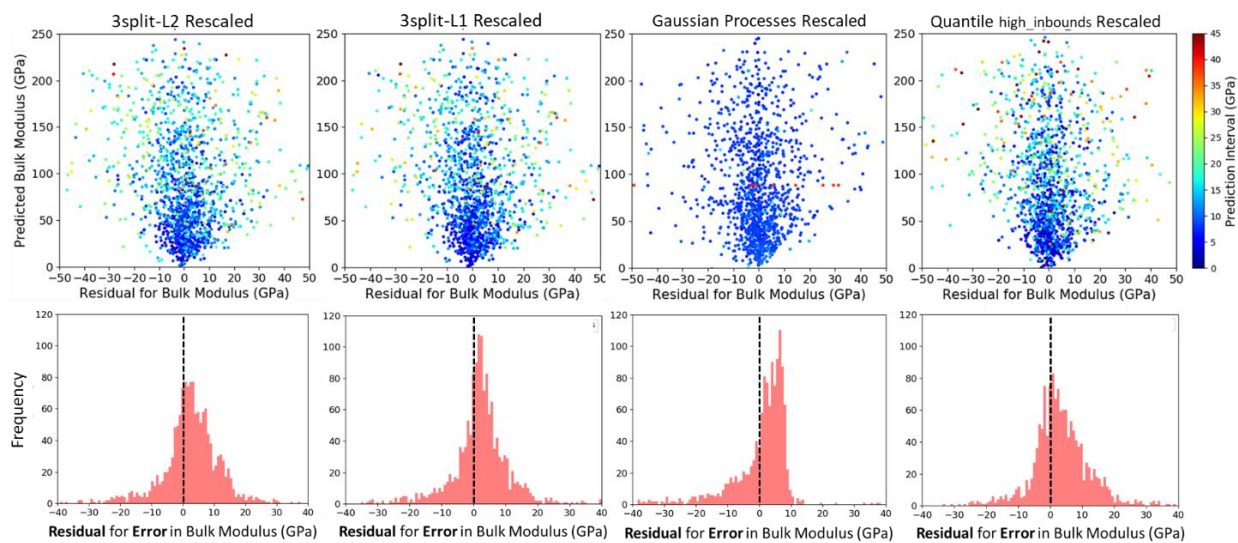

*Figure 6 Top row: Residual for base model (predicted property – DFT value) versus predicted property, in the case of Bulk Modulus, for the four examined approaches. The data point color indicates the predicted error (prediction interval). The distribution of predicted intervals depends on the modeling approach. Bottom raw: histograms for error model residual (predicted error – exact error). An even stronger dependence on the modeling approach is evident here, as the height, range and symmetry of the residual distribution vary significantly among error models.*

MAEs give important information on how a model behaves, but not complete. In Fig. 6 the residual for the base property (top row) and for the error prediction (bottom row) are displayed, in the case of Bulk Modulus. Similar results are found for all the other properties investigated in this work. In the top row images, the data point color indicates the predicted error (i.e., prediction interval). These results show that, while all the approaches are reasonably accurate both as base (most data falls between +10 GPa and -10GPa) and as error model (most of blue datapoints fall between +10 GPa and -10GPa and non-blue data fall outside this range), the dispersion of the error distribution does depend on the modeling approach. In other words, how many blue points actually fall outside the ±10 GPa range and, conversely, how many not-blue points fall inside such range, does depend on how the prediction interval is determined. To directly investigate the uncertainty in the error prediction, we plot the residual histogram for the prediction intervals models (Fig. 6, bottom row). A strong dependence on the modeling approach is evident from these plots, larger than the similarities among MAEs could have led us to expect, as the height, range and symmetry of the residual distribution vary significantly among error models. Once again, 3split-L1 produces the best results, as its error-residual histogram is the most symmetric, is centered very close to zero, and is the tallest and the one with smallest full-width-half-maximum. As the base model is the same among 3split-L1 and 3split-L2, all the differences between these two cases are to be attributed to the different definition of "error". As already observed when commenting predicted values versus DFT ones for GP (Fig. 4b), GP is characterized by getting a smaller spread in the predicted uncertainties than the other models, which leads to overestimating the error of well predicted data and, in a few cases, to underestimate the error of badly predicted ones. This is evident by its histogram peak not being centered around zero, as very few prediction intervals are predicted to be almost zero, even when the corresponding property values are very close to the exact ones. Conversely, not many large overpredictions occur, as made evident by the sharp dropping off of the histogram after 10 GPa.

## 4. CONCLUSIONS

In this work, we compare 3 different approaches to determine the uncertainty on individual machine learning predictions (prediction intervals). Specifically, we probe the Quantile loss function approach, machine learning the error directly, where the error is defined either as the absolute difference (3split-L1) or the square of difference (3split-L2) between the predicted and the observed values, and using Gaussian Processes. All approaches are applied to the modeling of 12 physical properties, ranging from energetic-related ones to elastic, optical, electronic and more. This investigation is necessary because quantities like MAE only evaluate ML models in a statistical manner, while users of such models need to know how reliable the specific prediction, they are interested in is. We identify each approach advantages and disadvantages, and we learned that descriptor choice matters. With good descriptors (like the CFID), even ML algorithms with default hyperparameters (Random Forest, in our investigation) give acceptable results. The choice of hyperparameters is particularly crucial when developing Gaussian Processes models. We find that Gaussian Processes give a good estimate for prediction intervals, although a bit overestimated, but are more time consuming to fit that the other approaches. Using the Quantile approach requires fitting 3 models and, in general, gives lower inbounds results than Gaussian Processes or 3split-L2. However, because of the 3 models, it is the easiest approach to fit for specific inbounds results. Machine learning the error directly has the enormous advantage of allowing the use of any loss function. However, it requires splitting the dataset in three parts, which could be a problem if the dataset is small to begin with. How the error is modelled also makes a difference and using the absolute difference between predicted and expected (3split-L1) was found to be, overall, the best approach. All data for training and testing were taken from the publicly available JARVIS-DFT (https://jarvis.nist.gov/jarvisbdft/ ) database, and the codes developed for computing the prediction intervals are available through JARVIS-Tools (https://github.com/usnistgov/jarvis) GitHub.

# Supplementary Information Uncertainty Prediction for Machine Learning Models of Material Properties

Francesca Tavazza, Brian De Cost, Kamal Choudhary

Materials Science and Engineering Division, National Institute of Standards and Technology, Gaithersburg, MD, 20899, USA

Table 1  MAEs for base model, MAEs for error models and inbounds percentages for data in Fig. 5 a) and b)

| MAE BASE model | Form. Energy | Exfoliation Energy | e_hull | OPT gap | MBJ gap | bulk modulus | shear modulus | OPT epsx | MBJ epsx | spillage | Max Ir |
|---|---|---|---|---|---|---|---|---|---|---|---|
|  | eV/atom | eV/atom | eV | eV | eV | GPa | Gpa |  |  |  | cm^(-1) |
| Quantile | 0.12 | 42.51 | 0.050 | 0.26 | 0.50 | 7.95 | 8.16 | 0.63 | 0.59 | 0.33 | 63.53 |
| 3split-L2 | 0.15 | 37.62 | 0.098 | 0.29 | 0.59 | 9.53 | 9.80 | 0.70 | 0.64 |  | 80.25 |
| 3split-L1 | 0.15 | 37.38 | 0.057 | 0.29 | 0.59 | 9.49 | 9.52 | 0.70 | 0.65 |  | 77.15 |
| GP | 0.16 | 36.16 |  |  | 0.52 | 8.72 |  |  |  | 0.32 | 67.60 |
|  |  |  |  |  |  |  |  |  |  |  |  |
| MAE ERROR model | Form. Energy | Exfoliation Energy | e_hull | OPT gap | MBJ gap | bulk modulus | shear modulus | OPT epsx | MBJ epsx | spillage | Max Ir |
|  | eV/atom | eV/atom | eV | eV | eV | GPa | Gpa |  |  |  | cm^(-1) |
| Quantile | 0.11 | 26.67 | 0.039 | 0.17 | 0.49 | 7.08 | 7.06 | 0.50 | 0.42 | 0.21 | 61.50 |
| 3split-L2 | 0.12 | 48.26 | 0.106 | 0.26 | 0.51 | 7.54 | 7.64 | 0.58 | 0.62 |  | 66.62 |
| 3split-L1 | 0.09 | 35.17 | 0.038 | 0.17 | 0.397 | 5.87 | 6.07 | 0.45 | 0.44 |  | 44.39 |
| GP. | 0.15 | 45.79 |  |  | 0.60 | 9.82 |  |  |  | 0.27 | 66.88 |
|  |  |  |  |  |  |  |  |  |  |  |  |
| INBOUNDS PERCENTAGE | Form. Energy | Exfoliation Energy | e_hull | OPT gap | MBJ gap | bulk modulus | shear modulus | OPT epsx | MBJ epsx | spillage | Max Ir |

| | | | | | | | | | | |
|---|---|---|---|---|---|---|---|---|---|---|
| *Quantile* | *64.50* | *53.75* | *63.10* | *53.24* | *70.39* | *67.33* | *68.78* | *67.41* | *53.56* | *63.98* | *61.15* |
| *3split-L2* | *71.08* | *79.01* | *77.43* | *82.13* | *75.95* | *71.60* | *73.80* | *75.90* | *77.13* | | *67.86* |
| *3split-L1* | *58.28* | *80.24* | *64.65* | *61.83* | *61.617* | *59.67* | *61.92* | *62.42* | *61.49* | | *57.96* |
| *GP.* | *80.37* | *85.19* | | | *81.78* | *81.96* | | | | *78.22* | *83.18* |

Table 2 Scaling factors to achieve 68% inbounds count for all models/properties

| SCALING FACTOR | Form. Energy | Exf. Energy | e_hull | OPT gap | MBJ gap | Bulk modulus | Shear modulus | OPT epsx | MBJ epsx | spillage | Max Ir |
|---|---|---|---|---|---|---|---|---|---|---|---|
| *Quantile* | *1.12* | *1.60* | *1.150* | *1.52* | *0.93* | *1.00* | *0.04* | *1.00* | *1.48* | *1.10* | *1.25* |
| *3split-L2* | *0.93* | *0.80* | *0.70* | *0.58* | *0.83* | *0.95* | *0.92* | *0.82* | *0.76* | | *0.94* |
| *3split-L1* | *1.20* | *0.76* | *1.09* | *1.15* | *1.2* | *1.16* | *1.14* | *1.13* | *1.15* | | *1.20* |
| *GP* | *0.68* | *0.68* | | | *0.62* | *0.60* | | | | *0.76* | *0.61* |